\documentclass[runningheads]{llncs}
\usepackage{booktabs}
\usepackage{graphicx}
\title{ACE-Net: Biomedical Image Segmentation with Augmented Contracting and Expansive Paths}
\titlerunning{ACE-Net: Biomedical Image Segmentation}
\author{Yanhao Zhu\inst{1}\thanks{This work was performed at Institute of Automation, Chinese Academy of Sciences}
\and Zhineng Chen\inst{2}
\and Shuai Zhao\inst{2}
\and Hongtao Xie\inst{3}
\and Wenming Guo\inst{1}
\and Yongdong Zhang\inst{3}}
\institute{Beijing University of Posts and Telecommunications, Beijing, China \and Institute of Automation, Chinese Academy of Sciences, Beijing, China \\ \email{
  zhineng.chen@ia.ac.cn} \and University of Science and Technology of China, Hefei, China}
\authorrunning{Y. Zhu et al.}
\begin{document}
\maketitle
\begin{abstract}
Nowadays U-net-like FCNs predominate various biomedical image segmentation applications and attain promising performance, largely due to their elegant architectures, e.g., symmetric contracting and expansive paths as well as lateral skip-connections. It remains a research direction to devise novel architectures to further benefit the segmentation. In this paper, we develop an ACE-net that aims to enhance the feature representation and utilization by augmenting the contracting and expansive paths. In particular, we augment the paths by the recently proposed advanced techniques including ASPP, dense connection and deep supervision mechanisms, and novel connections such as directly connecting the raw image to the expansive side. With these augmentations, ACE-net can utilize features from multiple sources, scales and reception fields to segment while still maintains a relative simple architecture. Experiments on two typical biomedical segmentation tasks validate its effectiveness, where highly competitive results are obtained in both tasks while ACE-net still runs fast at inference.

\end{abstract}
\section{Introduction}
Biomedical image segmentation aims at partitioning an image into multiple regions corresponding to anatomical objects of interest. It is an essential technique that allows the quantification of shape related clinical parameters. Recently fully convolution networks (FCNs) have shown compelling advantages compared with traditional methods like Otsu thresholding, watershed segmentation, etc.

Among the developed architectures in this field, U-net \cite{Ronneberger2017U} is probably the most famous one. It consists of a contracting path and a symmetric expansive path that enable multi-scale feature extraction and successive feature aggregation, respectively. Moreover, there are skip-connections from contracting to expansive path at each scale to compensate the loss caused by downsampling. They are important for biomedical images whose differences are typically subtle \cite{Drozdzal2016The}. In various biomedical segmentation tasks, U-net outperforms several FCN architectures that achieve promising results on natural image segmentation thanks to this elegantly designed architecture.

Following the success of U-net, a number of research efforts have been devoted to further enhance the segmentation performance. We can broadly categorize them into three types. The first one emphasizes on extracting more discriminative features by employing advanced deep modules. For example, residual and dense blocks were leveraged by FusionNet \cite{Quan2016FusionNet} and FC-DenseNet \cite{jegou2017one}. The second type is deep supervision that utilizes the groundtruth to guide model learning at intermediate layers of a CNN. It yielded better performance on several segmentation tasks, as demonstrated in \cite{hu2018retinal}\cite{Xie2015Holistically}\cite{Zhang2018Deep}. The third type is devising architectures that stack or concatenate multiple U-net-like FCNs. For example, cascade-FCN \cite{christ2017automatic} was proposed for liver and tumor segmentation. It consisted of two U-nets. The first U-net aimed to segment liver from a CT slice while the second learned to segment lesions from the liver mask given by the first U-net. Zeng et al. \cite{zeng2019ric} also developed the RIC-Unet that segmented the contour mask and foreground of nuclei by using different branches. Despite satisfactory performance attained, these architectures are typically dataset-specific thus need to be modified when migrating to other tasks.

Standing on the shoulders of these studies, in this paper we develop a novel ACE-Net that aims to improve the performance of general biomedical segmentation. ACE-net inherits an u-shape architecture. It is also featured by applying the recently proposed ASPP module \cite{Chen2017Rethinking} to strengthen the contrasting path, which robustly extracts features covering multiple scales and reception fields. Meanwhile, the expansive path receives information from multiple sources including the raw image, features combined with detailed and global context from both the contrasting and expansive sides through different ways. They are densely connected to generate a comprehensive feature representation that is critical to the identification of boundary pixels. In addition, deep supervision on side response \cite{Xie2015Holistically} is applied to both paths to further facilitate the segmentation. By leveraging the augmented paths, ACE-net can intensively utilize the features to segment while still maintains a relative simple architecture. We apply ACE-net to two typical biomedical segmentation tasks, i.e., the segmentation of neuronal structures in electron microscopic stacks (EM segmentation) \cite{arganda2015crowdsourcing} and the segmentation of digital retinal images for vessel extraction (DRIVE) \cite{Staal2004Ridge} which is quite different with the previous one. Experiments demonstrate that ACE-net achieves highly competitive results in both tasks while still runs fast at inference.






\section{Proposed Method}
\begin{figure}[th]
\includegraphics[scale=0.49]{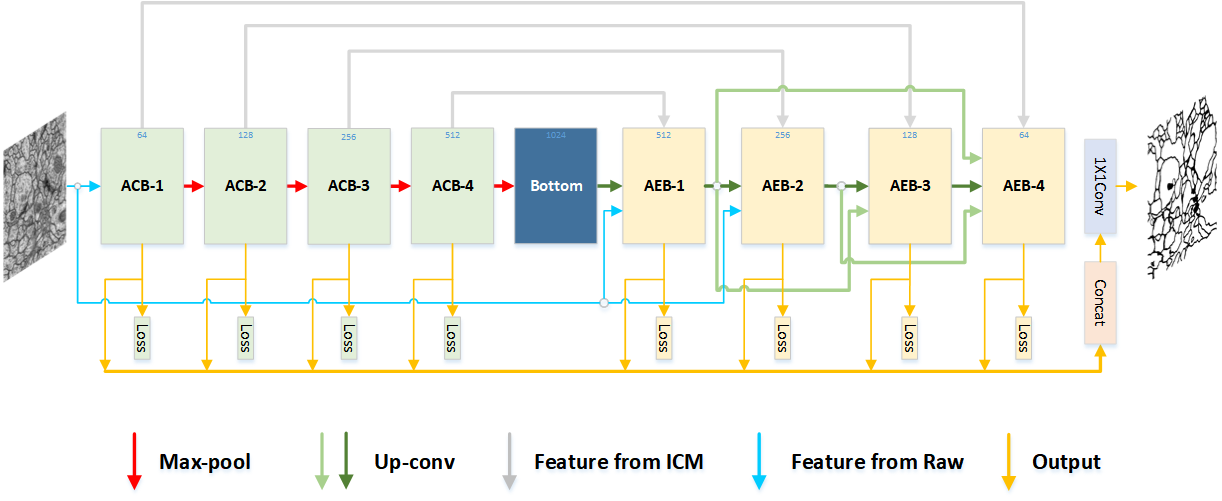}
\caption{An overview of the proposed ACE-net for biomedical image segmentation.}
\end{figure}
\noindent
The proposed ACE-net is presented in Fig.1, which is an augmented u-shape architecture with intensive connections among blocks. Specifically, on the contrasting path, four augmented contrasting blocks (ACBs) with size decreased progressively are stacked to extract discriminative features combined with context of multiple scales and reception fields. While on the expansive path, the same number of augmented expansive blocks (AEBs) are developed to integrate features from multiple sources to better localize. Besides, side output is applied to give deep supervision in every block.

\subsection{Augmented Contrasting Block with Intensive Context Modeling}

To effectively explore the feature locally and holistically, an intensive context modeling (ICM) module is introduced to augment the contrasting block (see Fig.2 (a)). In particular, besides the same contrasting block as in U-net, an ICM is added for feature enhancement. It separably takes as input the two 3x3 convolutions from the contrasting block, each first followed by another 3x3 convolution for feature abstraction and then concatenated. The generated feature maps are fed into a 1x1 convolution for dimension regularization.
Then, an ASPP module with different rates is carried out for context modeling at multiple reception fields. A 1x1 convolution is followed by and the resulted feature maps are concatenated with the corresponding expansive block via a skip-connection. Parallel to this, another 1x1 convolution is also applied for the purpose of dimension reduction, and a side output is calculated on this branch.

Two kinds of augmentation are obtained from the ACB. First, by employing ASPP, a block can model context at multiple reception fields rather than a single one as in U-net. It significantly enriches the feature passed to the expansive side and thus can alleviate the intra-class inconsistency problem which is elaborated in \cite{yu2018learning}. Second, deep supervision on the side branch is carried out. It is beneficial to extracting more targeted features and improving the segmentation.

\begin{figure}[h]
\begin{minipage}[b]{.5\linewidth}
  \includegraphics[scale=0.24]{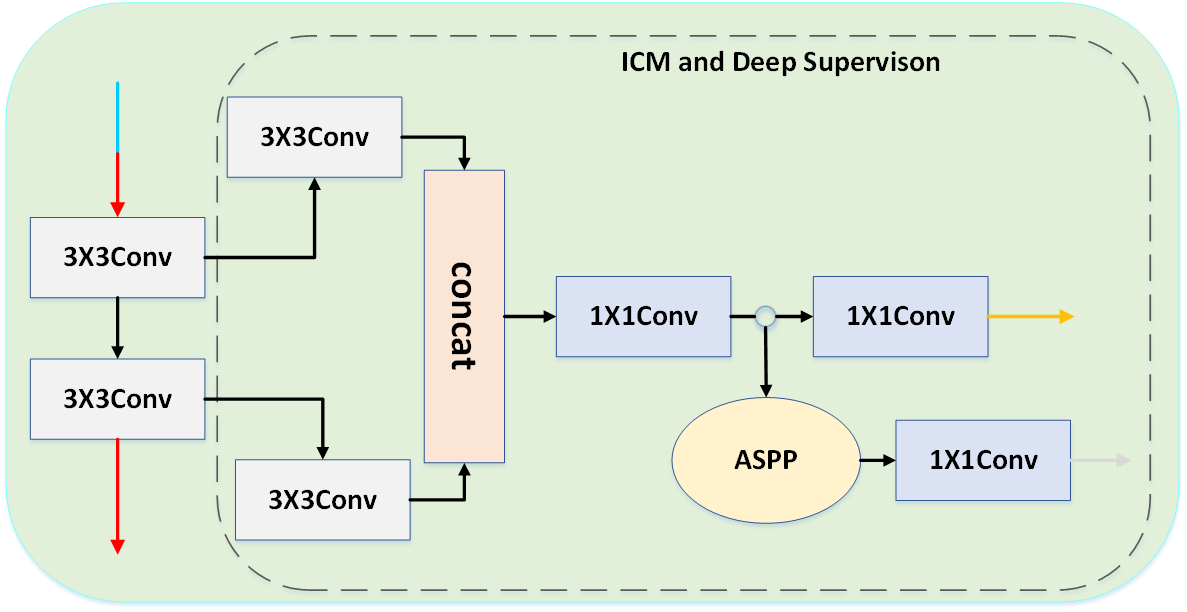}
  \centerline{\scriptsize (a) Augmented Contrasting Block}
\end{minipage}
\begin{minipage}[b]{.5\linewidth}
  \includegraphics[scale=0.24]{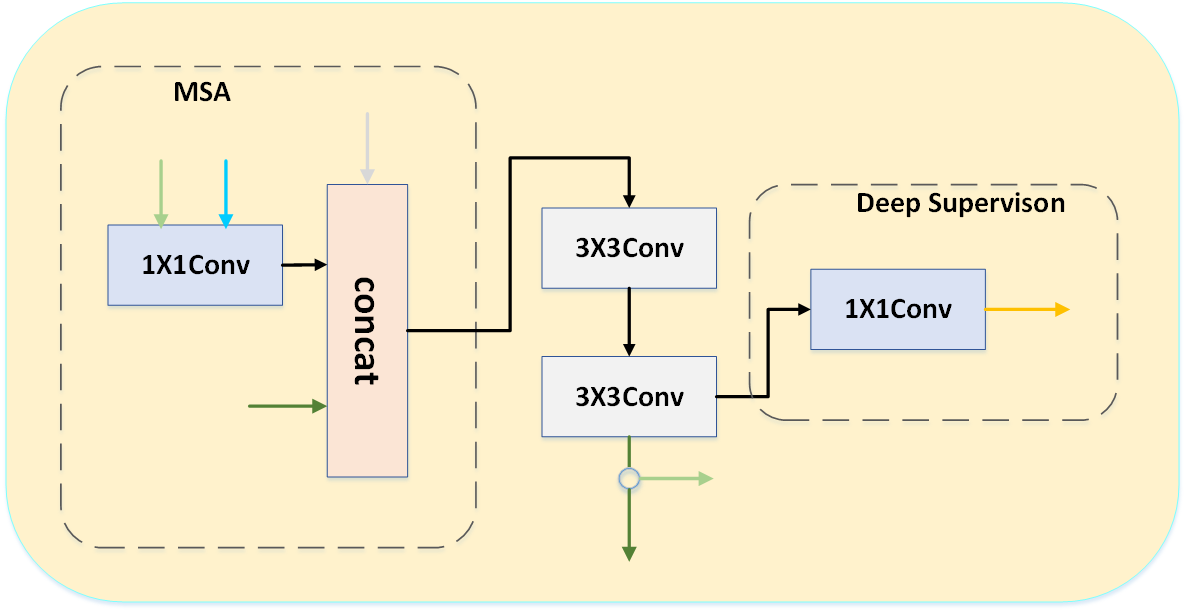}
  \centerline{\scriptsize (b) Augmented Expansive Block}
\end{minipage}
\caption{ Detail structures of (a) Augmented Contrasting Block and (b) Augmented Expansive Block in the proposed ACE-net.}
\end{figure}


\subsection{Multi-Source Aggregation for Augmented Expansive Block}
We also devise a multi-source aggregation (MSA) module to augment the expansive block (See Fig.2(b)). It is combined with the raw expansive block to generate a comprehensive feature representation for better segmentation. To be specific, let MSA-\emph{i} be the MSA of the \emph{i}-th AEB in the expansive path, it can concatenate up to four kinds of input from different sources: feature from the bottleneck layer (for \emph{i}=1) or previous AEB (for the rest), the \emph{i}-th ACB feature, the resized raw image, and densely connective features from previous AEBs except for its direct ancestor, e.g., AEB-1 and AEB-2 for MSA-4. Before feeding into the MSA, the latter two kinds of features experience a 1x1 convolution for dimension alignment, i.e., making different inputs have an equal number of feature maps. Then, two 3x3 convolutions the same as the raw expansive block are applied to generate features for the following stacked AEBs. Meanwhile, to establish deep supervision, the two 3x3 convolutions are followed by a 1x1 convolution, whose output is utilized to calculate the side loss. Note that the deep supervision in AEB undergoes less convolutions, as its features are deeper and more abstract compared with those in ACB.

The augmentation of AEB is threefold. First, it receives rich features from different sources covering multiple scales and reception fields, which are critical for resolving the challenging edge ambiguity. Second, explicit supervision from the raw image is obtained. It contains the most accurate location information and is helpful to the segmentation especially for those small-sized convolutional layers. It is also one peculiar feature of ACE-net as existing work seldom directly supervises in this way. Third, deep supervision is again carried out at every block, which serves as both learner and regularizer to the whole network.

\subsection{Loss Function}
Following U-net, ACE-net employs the sparse softmax cross-entropy as loss function, which is computed by a pixel-wise softmax over the feature maps combined with the cross-entropy. We calculate it for the final output and the eight side outputs. The overall loss function is given by
$$ L = L_p(\omega) + \lambda\sum_{n=1}^{M}L_s^{(n)}(\omega)$$
where \emph{M} is the total number of ACB and AEB, $L_p$ and $L_s$ are the final loss and the side loss, respectively, $\lambda$ is a hyper-parameter to balance $L_p$ and $L_s$. The same as \cite{Xie2015Holistically}, we set $\lambda=1$ in all experiments.



\section{Experiments}
We evaluate the proposed ACE-net on two longstanding but still active biomedical segmentation tasks: EM segmentation \cite{arganda2015crowdsourcing} and DRIVE \cite{Staal2004Ridge}. We first introduce our implementation details, followed by the ablation study and experimental results on EM segmentation. Finally, we show the result on DRIVE.

\subsection{Implementation Details}
We implemented ACE-net by using TensorFlow (version 1.4.0). Data augmentations including flip, zoom and rotate were employed. The network was trained on one NVIDIA TITAN Xp GPU with a mini-batch size of one. Adam optimizer was used to optimize the whole network and the learning rate was fixed to 1e-4. It took approximate 10h and 14h to train ACE-net from scratch on the EM segmentation and DRIVE datasets, respectively. Besides, ACE-net runs fast on both datasets at inference. It took 0.53s to segment a 512x512 EM neuronal image, and 0.83s to segment a 592x576x3 retinal image on average.

\subsection{Experiments on EM Segmentation}
The segmentation of neural EM images is an important task in observing the structure and function of brain.
The EM segmentation dataset has 30 training images with publicly available groundtruth and 30 test images with annotations kept private for the assessment of segmentation accuracy. Website of the task remains open for submission although the competition is in year 2012. The evaluation metric is $Vrand$, which is elaborated in \cite{arganda2015crowdsourcing}.
\subsubsection{Ablation Study}
\noindent
\begin{table}[h]
\centering
\caption{Performance of ACE-net under different settings. CP(\textbf{C}): the contrasting path. EP(\textbf{E}): the expansive path. ICM(\textbf{I}): the intensive context modeling module. MSA(\textbf{M}): the multi-source aggregation module. DS(\textbf{D}): deep supervision.}
\makeatletter\def\@captype{table}
\setlength{\tabcolsep}{3mm}{
    \begin{tabular}{lc}
    \toprule
    Method  & $Vrand$ \\
    \midrule
    C+E+I+D& 0.9725 \\
    C+E+I+D+M(1)& 0.9774 \\
    C+E+I+D+M(1,2)&\textbf{0.9797}\\
    C+E+I+D+M(1,2,3)& 0.9746 \\
    C+E+I+D+M(1,2,3,4)& 0.9726 \\
    \bottomrule
    \end{tabular}}%
\makeatletter\def\@captype{table}
    \setlength{\tabcolsep}{3mm}{
    \begin{tabular}{lc}
    \toprule
    Method  & $Vrand$ \\
    \midrule
    C+E+M+D & 0.9723 \\
    C+E+M+D+I(1)  &0.9730 \\
    C+E+M+D+I(1,2) & 0.9760 \\
    C+E+M+D+I(1,2,3) & 0.9769 \\
    C+E+M+D+I(1,2,3,4) & \textbf{0.9797}\\
    \bottomrule
    \end{tabular}}%
\end{table}
To better understand ACE-net, we execute controlled experiments on the EM dataset. All the experiments are under the same setting  except for the specified difference.
\begin{figure}[h]
\centering
\begin{minipage}[b]{.24\linewidth}
  \includegraphics[scale=0.22]{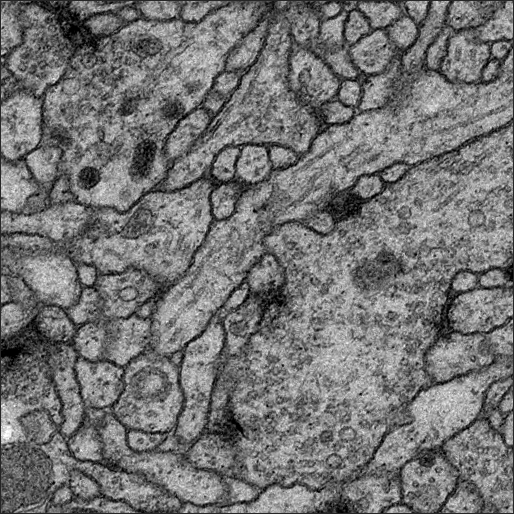}
\end{minipage}
\begin{minipage}[b]{0.24\linewidth}
  \includegraphics[scale=0.22]{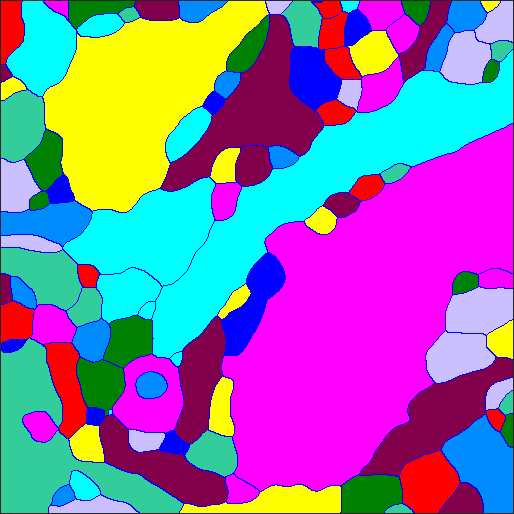}
\end{minipage}
\begin{minipage}[b]{.24\linewidth}
  \includegraphics[scale=0.22]{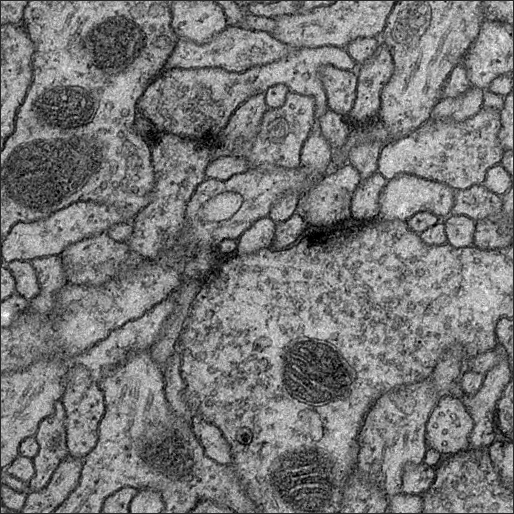}
\end{minipage}
\begin{minipage}[b]{0.24\linewidth}
  \includegraphics[scale=0.22]{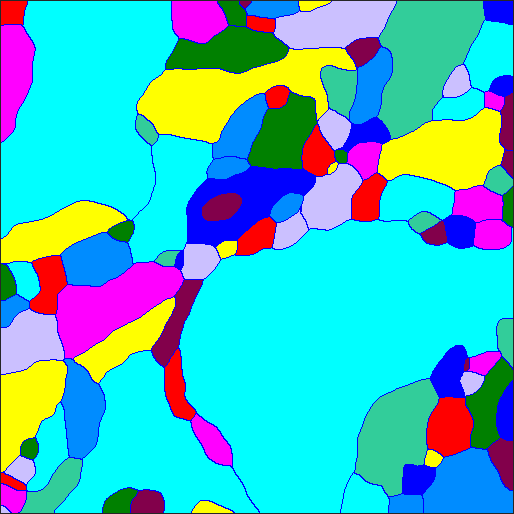}
\end{minipage}
\caption{Segmentation instances on ISBI 2012 EM Segmentation (slice 7 and 30).}
\label{figure:res}
\end{figure}

As explained, the ICM and MSA modules are important w.r.t ACE-net. To assess their effects, we first equip all the ACBs with the ICM module and stepwisely add the MSA module to AEB. Specifically, we test the absence/existance of the resized raw image only, as the effectiveness of skip-connections and dense connections have been proven in \cite{Drozdzal2016The}\cite{bilinski2018dense}. As shown in Tab.1, the best result is obtained when the raw image is available for the first two MSAs. This is reasonable as the subsequent AEBs are connected with shallow ACBs, which convey rich location information such that supplement from the raw image is less meaningful. We then keep the raw image seen by the first two AEBs and evaluate the ICM module. It is seen that the incorporation of all the four ICM modules yields the best performance, indicating the effectiveness of applying ASPP to extract features. In view of the study above, we determine the final architecture of ACE-net, as depicted in Fig.1.



\subsubsection{Results}

We submit the result of ACE-net to the website of EM segmentation, and list several top-ranked results and their papers as well as ours. The challenge lasted seven years and the leader board accumulated over 180 different submissions. From Tab.2, we can seen that a performance gain of 0.007 was obtained when compared U-net with ACE-net without post-processing, which yielded a prominent difference of over 30 rankings in the leader board. It indicates that ACE-net is suitable for biomedical segmentation. We also presented the results with the post-processing of \cite{beier2017multicut}, which were used by all the four methods listed in the right of Tab.2. Two images segmented by ACE-net are shown in Fig.3. Our result is highly competitive compared with the state-of-the-arts.


\begin{table}[htb!]
\centering
\caption{Results of published entries on the EM segmentation task \cite{arganda2015crowdsourcing}. The full leader board is available at: \emph{http://brainiac2.mit.edu/isbi\_challenge/leaders-board-new}}
\makeatletter\def\@captype{table}
\setlength{\tabcolsep}{2mm}{
    \begin{tabular}{lc}
    \toprule
    Method(without post-process)  & $Vrand$ \\
    \midrule
    ACE-net(ours) & \textbf{0.9797}\\
    M2FCN \cite{Shen2017Multi} & 0.9780 \\
    FusionNet \cite{Quan2016FusionNet}& 0.9780 \\
    U-net \cite{Ronneberger2017U}& 0.9727 \\
    \bottomrule
    \end{tabular}}%
\makeatletter\def\@captype{table}
    \setlength{\tabcolsep}{2mm}{
    \begin{tabular}{lc}
    \toprule
    Method(with post-process)  & $Vrand$ \\
    \midrule
    SFCNN \cite{Weiler2017Learning} & \textbf{0.9868}\\
    ACE-net(ours) & 0.9850 \\
    M2FCN \cite{Shen2017Multi}& 0.9838\\
    IAL LMC \cite{beier2017multicut} &0.9822 \\
    \bottomrule
    \end{tabular}}%

\end{table}
\subsection{Experiments on DRIVE}
\begin{figure}[h]
\centering
\begin{minipage}[b]{.238\linewidth}
  \includegraphics[scale=0.146]{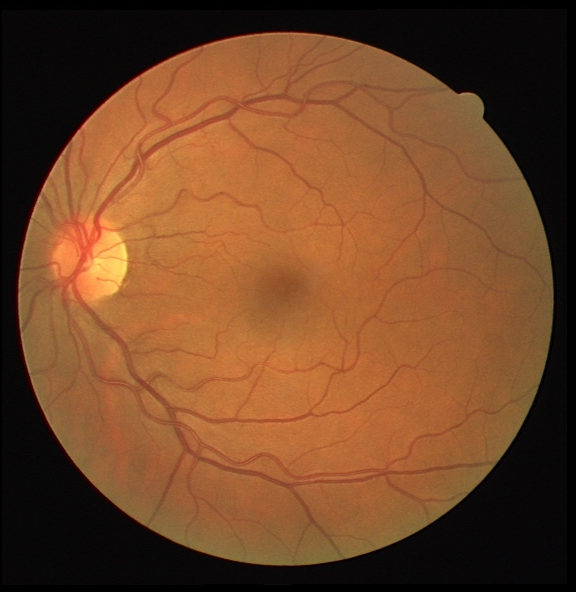}
\end{minipage}
\begin{minipage}[b]{0.238\linewidth}
  \includegraphics[scale=0.146]{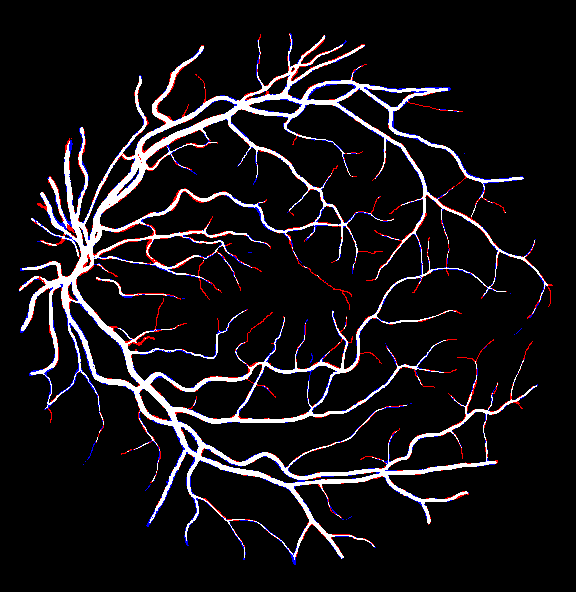}
\end{minipage}
\begin{minipage}[b]{.238\linewidth}
  \includegraphics[scale=0.146]{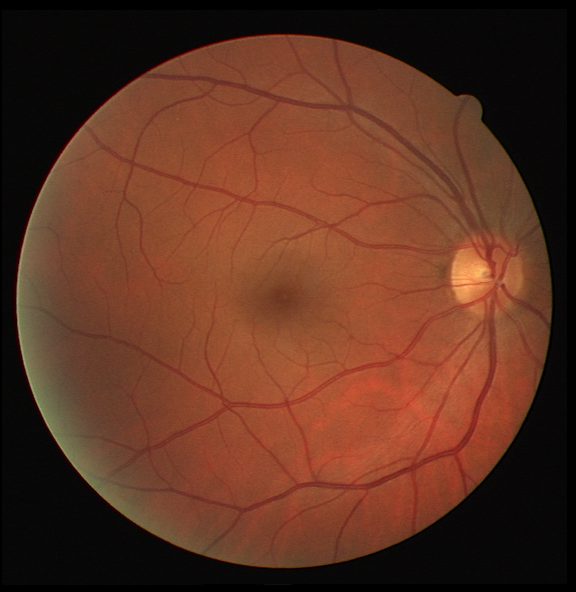}
\end{minipage}
\begin{minipage}[b]{0.238\linewidth}
  \includegraphics[scale=0.146]{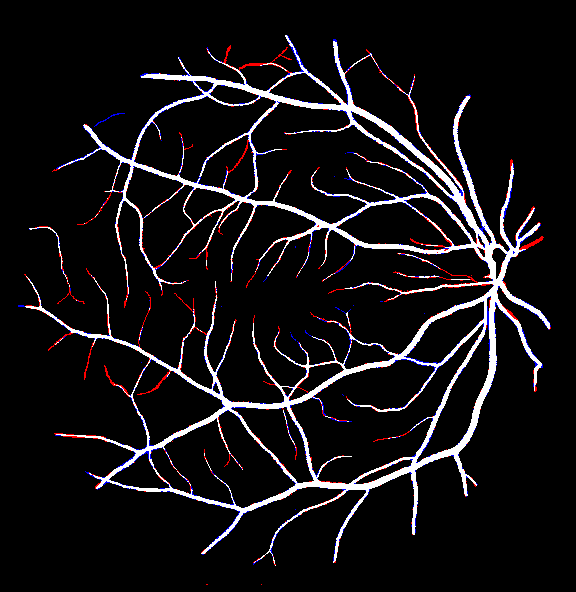}
\end{minipage}
\caption{Results on DRIVE (image 1 and 19). For the segmented images, correct detections are white, misses are red, and false positives are blue.}
\label{figure:res}
\end{figure}
The DRIVE dataset includes 40 color fundus photographs divided into two parts: 20 training images and 20 test images with manual segmentation of the vasculature and binary masks of FOV. As for the evaluation metrics, we use sensitivity, specificity, accuracy and the area under the ROC curve (AUC). The measurements are calculated only for pixels inside the FOV area. Tab.3 lists several top results as well as ours.


\begin{table}[h]
\caption{Comparison to start-of-the-art methods on DRIVE dataset.}
\centering
\setlength{\tabcolsep}{1.5mm}{
\begin{tabular}{lcccc}
\toprule
Method&Sensitivity(\%)&Specificity(\%)&Accuracy(\%)&AUC(\%)\\
\midrule
Azzopardi et al. \cite{Azzopardi2015Trainable}&76.55&97.04&94.42&96.14\\
Li et al. \cite{li2016cross} &75.69&98.16&95.27&97.38 \\
Orlando et al. \cite{Orlando2016A}  &78.97 &96.84 &N.A &95.07 \\
Dasgupta et al. \cite{dasgupta2017fully}  &76.91 &98.01 &95.33 &97.44 \\
Y.Wu et al. \cite{wu2018multiscale}  &78.44 &98.19 &95.67 &\textbf{98.07} \\
Y.Zhang et al. \cite{Zhang2018Deep} &\textbf{87.23} &96.18 &95.04 &97.99 \\
ACE-net(ours) &77.25 &\textbf{98.42} &\textbf{95.69} &97.42\\
\bottomrule
\end{tabular}}%
\end{table}

It is seen that ACE-net also gets competitive results on DRIVE without any modification in architecture or training settings. It attains the highest values on specificity and accuracy while the values for the other two metrics are still competitive except for the sensitivity of \cite{Zhang2018Deep}. The prominent higher sensitivity of \cite{Zhang2018Deep} is largely attributed to a dedicated pre-processing, which enhances the segmentation of the capillaries significantly. However, the pre-processing is network-independent. It also can be combined with ACE-net to seek for a better sensitivity although we have not implemented it now. Besides, ACE-Net performs similar to \cite{wu2018multiscale}, but it is likely to run 10x faster on a similar server due to a simpler architecture. Two instances segmented by ACE-net are shown in Fig.4.

\section{Conclusion}
Aiming at improving the performance on different biomedical segmentation applications, we devise ACE-net with augmented contracting and expansive paths, where advanced modules and novel connections are introduced to enhance the feature representation from multiple sources, scales and reception fields. The experiments conducted on EM segmentation and DRIVE datasets basically validate our proposal, where competitive results to state-of-the-arts are obtained in both tasks while ACE-net still maintains a fast inference speed.
\\\\
\noindent
\textbf{Acknowledgement.} This work is supported by the National Natural Science Foundation of China under grant no. 61772526

%
%
%
\bibliographystyle{splncs04}
\bibliography{paper1900}
\end{document}